\documentclass[sigconf,screen]{acmart}
\AtBeginDocument{%
  \providecommand\BibTeX{{%
    \normalfont B\kern-0.5em{\scshape i\kern-0.25em b}\kern-0.8em\TeX}}}

\definecolor{oegcolor}{rgb}{0.6,0.2,0.6}

\begin{document}

%\title{Facilitating Model Comparison Studies Through Reproducibility Factors}
%\title{Sources of Irreproducibility i Machine Learning: A Framework}
%\title{A Framework for Operationalizing the Scientific Method and Improving the  Reproducibility of Machine Learning Studies}
%\title{Improving the Evaluation of Model Comparison Studies Trough Operationalizing the Scientific Method and Reproducibility}
%\title{Improving the Evaluation of Model Comparison Studies Trough an Operationalization of the Scientific Method and Reproducibility}
%\title{Towards Improved Methodology for Machine Learning Trough an Operationalization of the Scientific Method and Reproducibility}
%\title{Operationalizing Machine Learning Reproducibility: A Framework with Practical Examples}
%Operationalizing Machine Learning Reproducibility: \\% How Design Choices Affect the Conclusion of Experiments
%How Conclusions Are Affected by Experiment Design Choices}
%\title[Operationalizing Machine Learning Reproducibility]{Operationalizing Machine Learning Reproducibility through a Categorization of Experiment Design Choices}

\title[Sources of Irreproducibility]{Sources of Irreproducibility in Machine Learning: A Review }

\author{Odd Erik Gundersen}
\email{odderik@ntnu.no}
\affiliation{%
  \institution{Norwegian University of Science and Technology}
  \city{Trondheim}
  \country{Norway}
}
\affiliation{%
    \institution{Aneo AS}
    \city{Trondheim}
    \country{Norway}
}

\author{Kevin Coakley}
\email{kcoakley@sdsc.edu}
\affiliation{%
  \institution{Norwegian University of Science and Technology}
  \city{Trondheim}
  \country{Norway}
}
\affiliation{%
  \institution{San Diego Supercomputer Center, UC San Diego}
  \city{La Jolla}
  \country{USA}}

\author{Christine R. Kirkpatrick}
\email{christine@sdsc.edu}
\affiliation{%
  \institution{San Diego Supercomputer Center, UC San Diego}
  \city{La Jolla}
  \country{USA}}

\author{Yolanda Gil}
\email{gil@isi.edu}
\affiliation{%
 \institution{Information Sciences Institute, USC}
 \city{Los Angeles}
 \country{USA}}

%\renewcommand{\shortauthors}{Gundersen, et al.}

\begin{abstract}
\textbf{Background:} 
Many published machine learning studies are irreproducible. 
Issues with methodology and not properly accounting for variation introduced by the algorithm themselves or their implementations are attributed as the main contributors to the irreproducibility.% of these studies.
\textbf{Problem:} 
There exist no theoretical framework that relates experiment design choices to potential effects on the conclusions.
Without such a framework, it is much harder for practitioners and researchers to evaluate experiment results and describe the limitations of experiments. %for both practitioners and researchers 
The lack of such a framework also makes it harder for independent researchers to systematically attribute the causes of failed reproducibility experiments.
\textbf{Objective:} 
The objective of this paper is to develop a framework that enable applied data science practitioners and researchers to understand which experiment design choices can lead to false findings and how and by this help in analyzing the conclusions of reproducibility experiments.
\textbf{Method:} 
We have compiled an extensive list of factors reported in the literature that can lead to machine learning studies being irreproducible.
These factors are organized and categorized in a reproducibility framework motivated by the stages of the scientific method.
The factors are analyzed for how they can affect the conclusions drawn from experiments.
A model comparison study is used as an example. 
\textbf{Conclusion:} 
We provide a framework that describes machine learning methodology from experimental design decisions to the conclusions inferred from them.

%?The reproducibility framework that we present here will support practitioners and researchers to analyze and understand experiment design decision, how they limit the generality of conclusions and evaluate how these can affect the conclusion of an experiment. 
%We provide a theoretical framework that supports researchers and practitioners in analyzing  how experiment design decisions can affect the conclusion of experiments and evaluate the limitations of the conclusions drawn from them.

%Progress in machine learning research is often demonstrated by model comparison studies that show improvements over the state-of-the-art using benchmark datasets.  Because reproducing published results can be difficult, authors are encouraged to follow “reproducibility checklists” that motivate the publication of data and software together with other experimental details.  However, reproducibility of model comparison studies remains challenging.  This paper presents an extensive compilation of reproducibility factors reported in the literature that can affect the outcome of model comparison experiments.  These factors are organized and categorized in a reproducibility framework motivated by the stages of the scientific method as well as varying degrees of reproducibility.  The paper also discusses the benefits to those reproducing prior studies in helping them identify the sources of differences in experiment outcomes, and paves the way for further methodological improvements in empirical machine learning.
\end{abstract}

\begin{CCSXML}
<ccs2012>
<concept>
<concept_id>10010147.10010257</concept_id>
<concept_desc>Computing methodologies~Machine learning</concept_desc>
<concept_significance>500</concept_significance>
</concept>
</ccs2012>
\end{CCSXML}

\ccsdesc[500]{Computing methodologies~Machine learning}

\keywords{Machine learning, research methodology, reproducibility, model comparison experiments}


\maketitle

\section{Introduction}
\label{sec:introduction}
% Define scope
% Make clear why the topic is important
% intended and existing applications
% Comprehensible with respect to the state-of-the-art
% Personal stance on the topic
% Description of existing work should be structured
% Should contain a discussion (identifying open questions and challenges)

%Clear wins are typically required for a research paper that describes a learning method to be published at a top venue \cite{sculley_2018}. %Se hvordan de formulerer clear wins
%The ImageNet Large Scale Visual Recognition Challenge \cite{Russakovsky_2015} is an example of a competition where the progress was accelerated through model comparison.
%Wagstaff \shortcite{wagstaff_2012} proposed to focus research on real world impact challenges such as improving the ELO rating of chess engines and discovering a new physical law using machine learning.
%Although several such impact challenges have been achieved lately, such as in game play \cite{silver_2016,mnih_2015}, protein folding, and software programming, model comparison studies are still the standard way of showing progress as such experiments are easier to conduct in a practical setting. 
%references: protein folding + software programming
%Lately, however, several model comparison studies that benchmark state-of-the-art models have shown that progress is not as steady as the impression given when one reads the scientific literature.
In recent years, many machine learning studies have shown to be very challenging to reproduce.
The areas of machine learning that have reported issues are very diverse and include forecasting \cite{makridakis_2018}, natural language processing \cite{belz_2021a}, generative adversarial networks \cite{lucic_2018}, deep reinforcement learning \cite{henderson_2018}, recommender systems \cite{dacrema_2019}, %Could be changed to 2021 reference if too many references
and image recognition \cite{bouthillier_2019}. 
%Their findings are that many state-of-the-art results published at top venues in machine learning are irreproducible, which adds to the reproducibility crisis \cite{hutson_2018}.
%Ioannidis \shortcite{ioannidis_2005} even argues that most research findings are false. 
%, and one of the causes of false findings in AI is that the research field is young and that analytical methods are still under experimentation.
The authors above point to many methodological issues that are commonly found in machine learning research.
Since applications of machine learning reach into many other fields \cite{gibney_2022}, methodological shortcomings can have far reaching effects particularly in domains with high-stakes decisions such as medicine \cite{roberts_2021,varoquaux_2022}, social sciences \cite{kapoor_2022}, psychology \cite{hullman_2022} and many more \cite{raji_2022}.

%%%%%%%%%%%%%% FIGURE SCIENTIFIC METHOD
\begin{figure*}[!t]
\centering\includegraphics[width=0.8\textwidth]{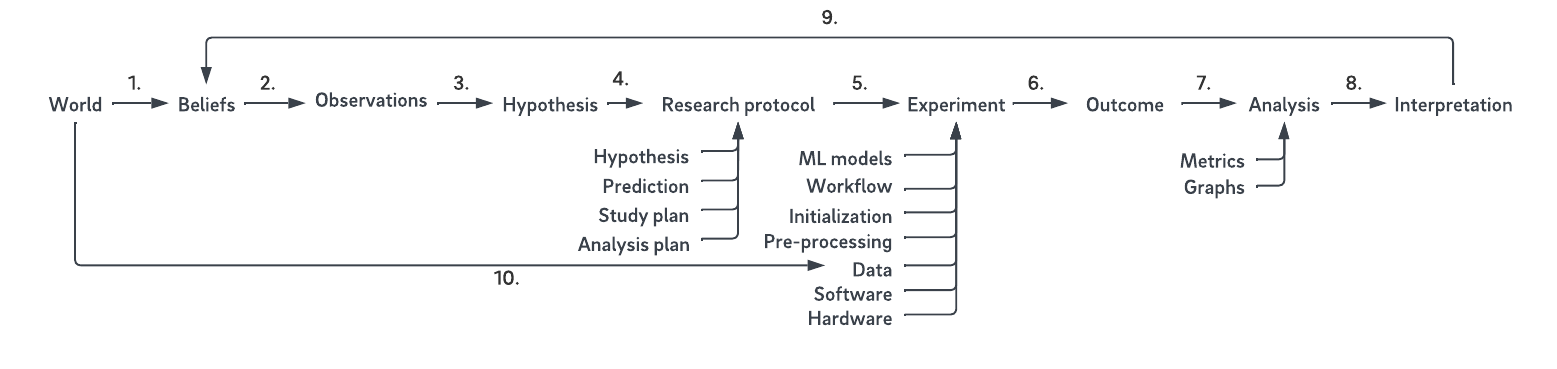}
\caption{
The \emph{scientific method} is a systematic process for acquiring knowledge about the world: 1) The world is observed, 2) explanations are made and testable statements are formulated, 3) experiments are designed to test the hypotheses and documented as research protocols, 4) experiments are implemented as code, 5) structured observations  are collected and stored digitally as data, 6) the implementation of the experiment is executed and outcomes are produced, 7) the outcomes are analyzed automatically by executing code, 8) the analysis is interpreted and a conclusion is reached to update beliefs \cite{gundersen_2021a}.
}
\label{fig:scientific_method}
\end{figure*}

Proper methodology requires a good understanding of which experiment design choices can lead to false findings. %irreproducible results. 
%Given the lack of awareness of how much machine learning results can vary, proper documentation that enables reproducibility becomes even more critical.
An experiment conducted by Pham et al. \shortcite{pham_2020} illustrated 16 identical training runs of a deep learning model that resulted in test accuracy varying from 9\% to 99\%.
%while implementation factors results in a 2.9\% difference.
The authors also presented a survey of 900 participants where 84\% were unsure or unaware about variance caused by how an experiment is implemented. 
Gundersen and Kjensmo \shortcite{gundersen_2018} found that experiments in AI presented at top conferences had incomplete documentation. 
The recognition of a reproducibility crisis in AI \cite{hutson_2018} has led to community-wide efforts to mitigate the crisis. 
The machine learning community is not alone in experiencing a reproducibility crisis \cite{button_2013,open_2015,baker_2016}. 

The machine learning and AI communities have introduced several mechanisms to improve the level of empirical rigor: 1) reproducibility checklists, 2) datasheets, 3) reproducibility challenges and 4) registered reports.  
Reproducibility checklists have been introduced at most top level machine learning and AI conferences, such as NeurIPS, AAAI, ICML, IJCAI, and EMNLP. 
Journals have still not made reproducibility checklists a default part of their submission procedure, although this is to change for JAIR (Journal of Artificial Intellgience Research) \cite{gundersen_2023}.
Datasets introduced at the NeurIPS Dataset and Benchmark Track are required to complete a data sheet \cite{gebru_2021} to ensure that  datasets are properly documented.
Reproducibility challenges have been introduced at both ICLR and NeurIPS \cite{pineau_2021} to encourage third-parties to try confirm findings of articles published at top-level conferences.
JAIR is to introduce reproducibility reports inspired by the reproducibility challenges \cite{gundersen_2023}. 
Finally, registered reports, which have been missing in AI and ML \cite{gundersen_2021b}, have been introduced at the journal ACM Transactions on Recommender Systems.

The framework presented here complements those efforts as it enables applied data science researchers and practitioners to: i) understand which experiment design choices can lead to false findings by providing an overview of design choices that can lead to irreproducible results, ii) understand how these design choices can affect the conclusion of experiments by mapping them to which part of the result elicitation process they belong to, iii) discuss limitations of experiments by using the overview of design decision and their consequences as a starting point for discussion, and iv) conduct and analyze reproducibility experiments by furthering their ability to understand and pinpoint the potential causes of failed reproducibility experiments.
Hence, the framework presented here could provide a methodological basis for reproducibility checklists and  provides justifications of why items should be reported.
The framework provides an opportunity to educate the community, as the items are not motivated.
For example, checklists tend to focus on the software side, while non-determinism must be controlled at all levels of the technical stack \cite{zhuang_2021}.
However, why is not clear to most researchers and practitioners according to \citet{zhuang_2021}.
Also, the design decisions presented in this paper extend existing reproducibility checklists significantly, by identifying many additional factors that need to be reported for a published experiment to be reproducible. %TODO: Add some numbers from Kevin's study?
Similarly, the framework could be useful when designing experiments for registered reports and as a justification for why data sheets are necessary. 

The main contribution of this paper is the identification and categorization of 41 design choices documented in the literature that can lead to false conclusions.
Another major contribution is a novel framework that enables applied data science researchers and practitioners to understand which experiment design choices can lead to false findings, understand how these design choices can affect the conclusion of experiments and conduct and analyze reproducibility experiments.
\section{A Reproducibility Framework}
\label{sec:model_comparison_and_reprodicibilty}

%%%%%%%%%%%%%% FIGURE SCIENTIFIC METHOD
%\begin{figure}[!t]
%\centering\includegraphics[width=0.5\textwidth]{figures/reproducibility_framework.pdf}
%\caption{
%TEST}
%\label{fig:scientific_method}
%\end{figure}

We follow the definitions proposed by \citet{gundersen_2021a}.
\emph{Reproducibility} is defined as the ability of independent investigators to draw the same conclusions from an experiment by following the documentation shared by the original investigators; a \emph{reproducibility experiment} is an experiment conducted by independent researchers to confirm the  findings of the original study using the documentation shared by the researchers that conducted the original study.
The documentation of a machine learning experiment is not restricted to written text in the form of a scientific report, it could also include the code and data. % used to conduct the experiment. 
However, additional documentation beyond a scientific report is not required for independent investigators to conduct a reproducibility experiment.
\citet{gundersen_2021a} specifies four different types of reproducibility studies based on which documentation is shared by the original researchers: \textit{R1 Description} if only textual documentation is shared, \textit{R2 Code} if text and code are shared, \textit{R3 Data} if text and data are shared and \textit{R4 Experiment} if text, data and code are shared.
Having access to code and data reduces the effort required to reproduce the results, while also leading to increased trust in the research \cite{gundersen_2019}. 
Drummond \shortcite{drummond_2009} argues that the power of a reproducibility experiment is greater with increased difference from the original study, which can be enforced by providing less documentation.
Still, code and data are commonly acknowledged as important for third parties to reproduce results \cite{haibe_2020}.
There are three degrees of reproducibility that are derived from the scientific method, which is illustrated in Figure \ref{fig:scientific_method}.
The idea will be exemplified by a model comparison study. 

Progress in machine learning is to a large degree driven by empirical evidence, and \emph{model comparison} is the standard method to identify the best performing machine learning model for a given task \cite{melis_2017,sculley_2018,bouthillier_2019,dacrema_2021}.
A \emph{model comparison study} is an experiment of which the objective is to decide which of a set of models has better performance.
Hence, a model comparison identifies the subset \textit{S} of a given set of computer programs \textit{C} that performs task \textit{T} better according to measure \textit{P} after learning from the same experience \textit{E}.
For a computer program, or model, to be considered a clear winner of a model comparison, it should be significantly better than the model that was previously considered state-of-the-art for the given task \cite{sculley_2018}. 
The hypothesis that a model is significantly better than the other models in the comparison should be tested statistically \cite{cohen_1995}. 
Empirical evidence is given by conducting experiments where the competing models learn from the same experiences under the same conditions. 

\textbf{Example:}
A reproducibility experiment is conducted to re-test the hypothesis that $m_{CNN}$ performs better than $m_{DNN}$ on the MNIST classification task as reported by \citet{lecun_1998}.
The documentation of the original study contained the paper cited above as well as the MNIST dataset that was famously published. 
No code was published, so a reproducibility study requires re-implementing the convolutional neural network and dense neural networks used in the original study.
Performance was measured using classification error and uncertainty was estimated, but how was not exactly described. 
As both the paper (text) and data (the MNIST dataset) are shared by the authors, and these resources are used for the reproducibility experiment, the reproducibility experiment is of type R3 Data.

\emph{Outcome reproducibility (O)} is achieved if the reproducibility experiment produces the same outcome as the original experiment.
The outcome of the image classification task is the set of labels given to each image in the test dataset.
If the model in the reproducibility experiment classifies each image with the exact same labels as they were assigned in the original experiment, the reproducibility experiment is outcome reproducible given that the analysis and conclusion of the original investigation are sound.
An experiment can only be evaluated for outcome reproducibility if the outcome produced in the original experiment is shared, which is the case for only 4\% of AI studies \cite{gundersen_2018}.
The outcome was not published by \citet{lecun_1998}, so our example reproducibility experiment could not be  outcome reproducible, but if it was, it would have been classified as \textit{OR3}.

\emph{Analysis reproducibility (A)} is achieved when the same analysis that was made by the original investigators lead to the same conclusion for the reproducibility experiment even if the outcomes differ.
Evaluating an experiment for analysis reproducibility requires the methods that were used to analyze the outcome to be shared.
%in the reproducibility experiment is different from the outcome produced by the original study, but the same analysis that was made by the original investigators lead to the same interpretation for both the original experiment and the reproducibility experiment. 
%In such as situation, the reproducibility experiment is analysis reproducible and supports the original hypothesis given that the same interpretation is done. 
Given our example, where code has to be re-implemented and executed in a different computing environment (we might not have access to an SGI server), the outcome will differ. 
However, as long as $m_{CNN}$ performs significantly better than $m_{DNN}$ in the reproducibility experiment when measuring the performance using error rate and uncertainty, the reproducibility experiment is analysis reproducible and would be classified as \textit{AR3}.

\emph{Interpretation reproducibility (I)} is achieved when a different analysis is done by independent investigators (on the same or different outcome) and their interpretation 
%, but the independent researcher’s interpretation 
of the analysis supports the conclusion drawn in the original experiment.
Hence, the conclusion of the original experiment is supported even though the reproducibility experiment produces different outcome and the outcome is analyzed in a different way i.e., by using the F1-score instead of the error rate, as long as the F1-score is significantly better for $m_{CNN}$ than for $m_{DNN}$. 
Evaluating for conclusion reproducibility requires that the methods used for analyzing the outcome are shared.
In our example, the reproducibility experiment would have been classified as \textit{IR3}.
%If the interpretation of an independent researcher when applying the same analysis on the same or similar outcome does not lead to the same conclusions, the reproducibility experiment does not support the original conclusion. 
%25.8\% of software engineering studies using DL share code and data \cite{liu_2020}
Figure

%%%%%%%%%%%%% SECTION - CATEGORIZING DESIGN DECISION
\section{Categorizing Design Decisions}
Many decisions about how to conduct and evaluate an experiment must be made before the experiment can be executed. 
Some of the decisions are made actively while others are made passively. 
In the end, all the decisions constitute the experiment, but some of these decisions can lead to changed outcomes,  analyses and interpretations of the analyses and thus false conclusions. 
These decisions are independent variables on which the experiment's conclusion depends.
Sometimes even seemingly small changes to these independent variables can lead to false conclusions.
Hence, these design decisions can be interpreted as potential sources of irreproducibility.

\begin{figure*}[!t]
\centering\includegraphics[width=1.0\textwidth]{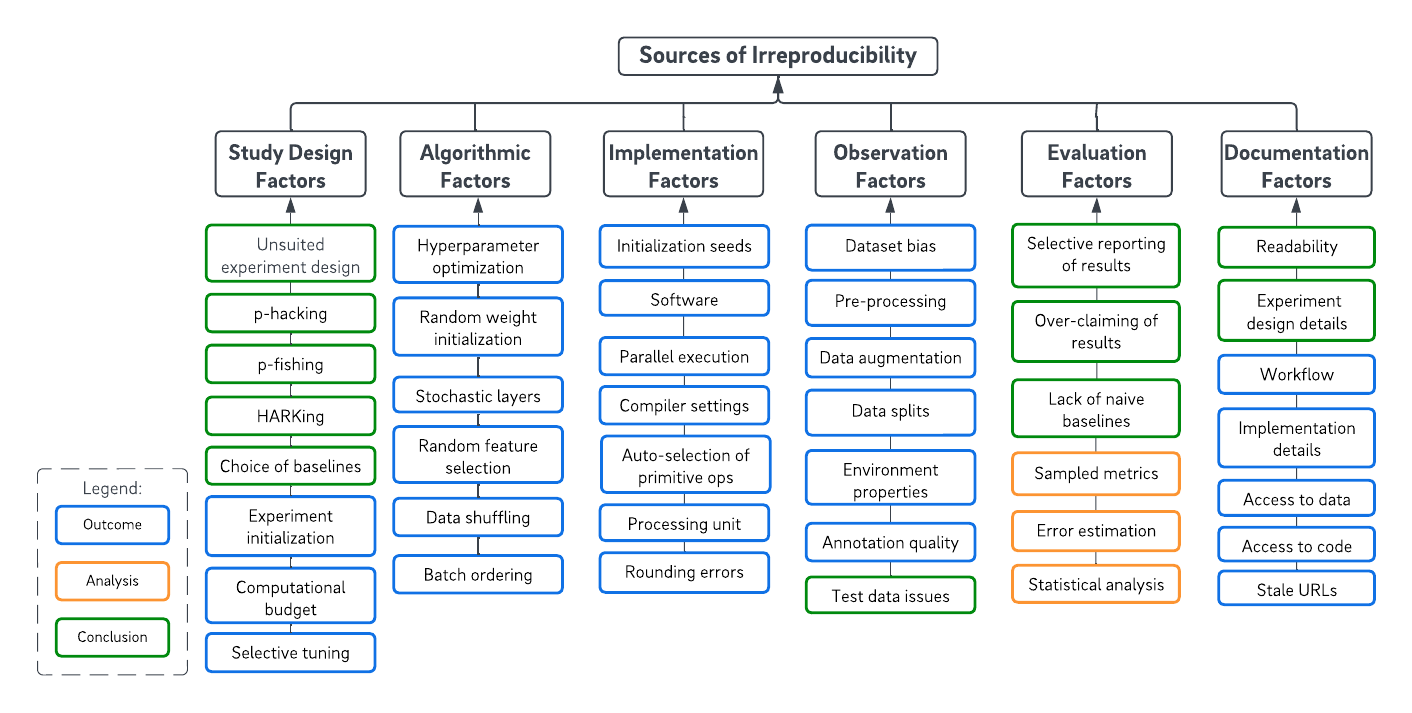}
\caption{Taxonomy of 41 design decisions and how they affect results, grouped into six categories. 
%\caption{Taxonomy of 45 key reproducibility factors that can affect outcome, analysis and interpretation of machine learning experiments, grouped into six major categories. 
%: 1) Study design factors, 2) Algorithmic factors, 3) Implementation factors, 4) Observational factors, 5) Evaluation factors and 6) documentation factors. 
}
\label{fig:taxonomy}
\end{figure*}

In this paper, we provide an overview of the design decisions found in the literature that can lead to false conclusions. 
We have identified 41 such design decisions and organized them into six major categories, shown in Figure \ref{fig:taxonomy}.  
These six categories, which we call factors in line with the terminology used by \citet{pham_2020}, comprise a super set containing the sources of variation that they introduced.
They group the sources of variation into algorithm-level and implementation-level factors,  
and they show how the sources of variation can change the outcome of a reproducibility experiment so much that the analyses lead to different conclusions if the variation is not controlled for.
Our approach is to provide an overview and taxonomy of design decisions that can lead to false conclusions and describe \textit{how} they can lead to false conclusions. 
We believe that this will not only be a valuable source for practitioners and researchers when designing experiments, but also when discussing limitations of conclusions and conducting reproducibility experiments. 

%%%%%%%%%%%%%FIGURE: TAXONOMY

%% Examples of how variables affect the outcomes after presenting the variables for each factorr?

%%%%%%%%%%%%% SECTION - STUDY DESIGN FACTORS
\section{Study Design Factors} %% FHva er factor: study design el. SoI?
Study design factors capture the decisions that goes into making the high-level plan for how to conduct and analyze an experiment in order to answer the stated hypothesis and research questions. 
%\begin{description}
%HERE!
    
    %%% Unsuited experiment design
    %\item 
    \textbf{Unsuited experiment design} These are experimental analyses that deviate from the claimed or implicit research goals \cite{cremonesi_2021}. 
    %, which do not affect the outcome. 
    Motivating why particular performance metrics, datasets or data pre-processing techniques are used should be given explicitly in order to ensure that the experiment design is suited \cite{dacrema_2021}.
    \textit{Doing analyses that do not support the research goals will likely lead to poor interpretations and thus the wrong conclusions.}
    
    %%% p-hacking
    %\item 
    \textbf{p-hacking} 
    When decisions of which data is included and which analysis is used are made during the analysis instead of in advance, researchers may self-servingly select the data and analysis that produce statistically significant results \cite{simonsohn_2014}.
    \textit{This will affect the interpretation and thus the conclusion.}
    
    %%% p-fishing
    %\item 
    \textbf{p-fishing} This term is used when seeking statistically significant results beyond the original hypothesis \cite{cockburn_2020}.
    \textit{Changing the hypothesis based on the p-value will not change the outcome nor the analysis, but the interpretation of the results will go beyond the original intent.}
    
    %%% HARKing
    %\item 
    \textbf{HARKing} (Hypotheses After Results are Known) is post-hoc reframing of experimental intentions to present a p-fished outcome as having been predicted from the start. 
    \textit{HARKing is to execute the scientific method backwards and will change the interpretation of the results, like p-fishing}.

    %%% Choice of baselines
    %\item 
    \textbf{Choice of baselines} For many machine learning tasks, it is often not clear what comprises the state-of-the-art. 
    Studies have found that many deep learning papers only compared against other deep learning algorithms, even though they were not performing better than other simpler baselines \cite{cremonesi_2021}.
    This could happen in cases where progress is shown by reusing experimental designs that propagate weak baselines without questioning them \cite{dacrema_2021}.
    \emph{Choosing a baseline that is inferior to the state-of-the art does not change the outcome of the target model, nor will it interfere with the analysis except that the baseline used for comparison is poorer than it could be. 
    The interpretation could change from a baseline doing better to target a model doing better.}%, which would change the conclusion.}
    
    %%% Experimental initialization
    %\item 
    \textbf{Experiment initialization} Differences in the setup or initialization of an experiment can lead to a difference between 5\% to 40\% in the number of solved instances of a SAT solver when running on the same hardware \cite{fichte_2021}, so they must be reported \cite{henderson_2018}.
    \textit{The setup can affect the outcome, so that the same analysis could be interpreted in a different way and change the conclusion if the variation under different setups is significant.}
    
    %%% Computational budget
    %\item 
    \textbf{Computational budget} Researchers with large computational budgets can sometimes prevent meaningful comparisons of algorithm performance experiments as they can spend the budget on intensive hyperparameter tuning of any given algorithm \cite{melis_2017,lucic_2018,dodge_2019,zhang_2020}.
    Bouthilier and Varoquaux \shortcite{bouthilier_2020} report that around 45\% of hyperparameters were manually tuned at NeurIPS 2019 and ICLR 2020, so systematic search could give a huge advantage. %Could argue for implementation instead
    %\textit{Larger computational budget could lead to more fair comparisons of algorithm performance.}
    \textit{Running algorithms for longer using more resources produces a different outcome.}
    
    %%% Tuning of algorithms
    %\item 
    \textbf{Selective tuning of algorithms} Researchers' favored algorithms are often fine-tuned to get the best possible result \cite{latifi_2021} while baselines are often not properly tuned \cite{cremonesi_2021,dacrema_2021}.  In some cases, old performance results are used to claim greater improvement than what could otherwise be claimed against the state-of-the-art \cite{crane_2018}. 
    %\textit{Tuning of algorithms in a selective, inconsistent manner could result in over-fitting and over-estimating and produce performing outcomes on test data that do not generalize well.}
    \textit{Tuning of algorithms in a selective, inconsistent manner could produce outcomes that do not reflect performance under the same conditions.}
%\end{description}

Study design factors can be controlled by designing a fair model comparison study where the hypothesis is tested genuinely and where the hypothesis is stated in such a way that the test properly answers the research question. 
A fair model comparison study assigns the same amount of resources, such as tuning and computational budget, to all models and sets up the experiment in a way that does not give advantages to a subset of the models or uses subpar baselines. 

%%%%%%%%%%%%% SECTION - ALGORITHMIC FACTORS
\section{Algorithmic Factors}
Algorithmic factors are design decisions to introduce stochasticity in the learning algorithms and training processes, 
%TODO: Provide examples + references of nice properties of randomness
%Randomness in training and inference 
leads to a different outcome for every experiment run. 

    %%% Hyperparameter optimization
    \textbf{Hyperparameter optimization} 
    Different hyperparameter optimization methods find different optimal hyperparameter values  \cite{reimers_2017,henderson_2018,bouthillier_2019,bouthillier_2021}, so researchers should specify exactly which method is used to improve reproducibility \cite{raff_2019,cooper_2021}. 
    Reimers and Gurevych \shortcite{reimers_2017} evaluated the variation of three different methods: random search, grid search and Bayesian optimization, and found that the variation that they caused is significant compared to other sources of variation.
    %{Specification of hyperparameter optimization methods greatly facilitates reproducibility. CK: changed not negligible to significant above}

    %%% Random weights initialization
    \textbf{Random weights initialization} 
    The initialization of weights in neural networks affects their performance \cite{pham_2020,zhuang_2021}. 
    %Typically, Gaussian distribution \cite{goodfellow_2016}. 
    Different initial weights might lead the hyperparameter optimization method to converge to local minima. 
    %{\em Randomization of weights initialization improves reproducibility}
    %TODO: CITE HERE% \cite{}. %several references in Pham %Sjekke bad seeds
    
    %%% Stochastic layers
    \textbf{Stochastic Layers} 
    Dropout, variational dropout, and noisy activations intended to make deep neural networks more robust end up affecting their performance \cite{reimers_2017,pham_2020,zhuang_2021}. 
    
    %%% Random feature selection 
    \textbf{Random feature selection} 
    Many learning algorithms rely on selecting features at random during training such as Random Forests \cite{breiman_2001}.
    %Features should be selected at random when needed for training, for example for Gradient Boosted Tree learning algorithms \cite{pouchard_2020}.
    The exact set of features selected will affect the outcome, and some selections might perform better than others \cite{pouchard_2020}.

    %%% Data shuffling
    \textbf{Data Shuffling} 
    Data samples are often shuffled randomly so that learning converges faster, which results in differences in outcome  \cite{reimers_2017,pham_2020,zhuang_2021}. 
    
    %%% Batch ordering
    \textbf{Batch ordering} 
    Because of memory limitations, data samples are fed into deep learning algorithms in batches. 
    Randomizing batch order between epochs results in different outcomes between training runs \cite{pham_2020,bouthillier_2021}. 

\textit{Relying on stochasticity cause outcomes to differ between experiment runs unless explicitly controlled for.}
Different combinations of initialization, training algorithm and dataset will lead to different outcomes that will perform differently.
If particularly lucky or unlucky, one might encounter single runs that might perform  very differently, even to such a degree that the results might affect the findings.
Algorithmic factors can be controlled by setting the pseudo-random number generator initialization seeds so that the outcome will be the exact same for each experiment run if everything else remains the same.
However, producing the same outcome over all runs does not mean that a finding is robust and generalizable.
Hence, the variation in the performance measured for the outcome produced over several experiment runs must be reported. 
As pointed out by Miller and Miller \shortcite{miller_2018}: \emph{"No quantitative results are of any value unless they are accompanied by some estimate of the errors inherent in them."}

%%%%%%%%%%%%% SECTION - IMPLEMENTATION FACTORS
\section{Implementation Factors}
Implementation factors are design choices related to the software and hardware that are used to execute the experiment.
These factors mirror the variations in physical sciences experiments that are introduced by conducting the same experiment in different laboratories.

    %%% Initialization seeds
    \textbf{Initialization seeds} Difference in the seeds used to initialize the pseudo-random number generator leads to difference in outcome \cite{melis_2017,bouthillier_2019}. 
    Reimers and Gurevych \shortcite{reimers_2017} show that the seed value for the random number generator can result in statistically significant ($p<10^{-4}$) results for different state-of-the-art systems.
    The same seed on different platforms produces different results \cite{pouchard_2020,gundersen_2022,nagarajan_2019}.
    %Averaging results over multiple runs using different seeds provides insight into population distribution.
    %Gundersen et al. \shortcite{gundersen_2022} pointed out that fixing seeds was not straightforward as they had to be fixed for many different libraries and that a bug in a framework prevented control of all stochasticity.
    
    %%% Software
    \textbf{Software} Outcomes across implementations of similar algorithms can vary significantly, e.g., TensorFlow vs. PyTorch \cite{pouchard_2020,henderson_2018}. 
    Hong et al. \shortcite{hong_2013} showed that a difference in operating systems affected the result. 
    %TODO: Revisit Hong: Is Hong's "software system" the same as Operating System?
    %%% Software version
    %\item \textbf{Software version} 
    Software versions  \cite{crane_2018,gundersen_2022,shahriari_2022},
    %%% Bugs in software
    %\item \textbf{Bugs in software} 
    bugs in either one's own implementation %of novel algorithms or baselines, 
    or  libraries, frameworks or operating systems might affect the outcomes \cite{crane_2018,pham_2020,pineau_2021,gundersen_2022}.
    
    %%% Non-deterministic ordering of floating point operations
    %\item \textbf{Non-deterministic ordering of floating-point operations}  
    
    %%% Parallel execution
    \textbf{Parallel execution} Random completion order of parallel tasks introduces variation \cite{pham_2020}. 
    Increased parallelism is a driver for noise \cite{zhuang_2021}. 
    Truncation error of floating point calculations introduces variability as $A+B+C \ne C+B+A$, when calculated in parallel \cite{pham_2020,gundersen_2022}. 
    Truncation error can be reduced but not completely removed by changing from single precision (32 bits) to double precision (64 bits) at a cost of doubling memory requirements and tripling the training time \cite{pinto_2021}.
    
    %%% Compiler settings
    \textbf{Compiler settings} Hong et al, \shortcite{hong_2013} found severe sensitivity to Intel compiler optimization levels %and found simulation results to be the same between O3 and O4 optimization levels but found the results to be different with each optimization level less than O3 
    for weather simulations that rely on huge amounts of floating point calculations, which is the case for machine learning too.
    
    %%% Auto-selection of primitive operations
    \textbf{Auto-selection of primitive operations} High level libraries implement deep learning algorithms using GPU-optimized deep learning primitives provided by low-level libraries such as cuDNN and CUDA \cite{pham_2020}. Autotune in cuDNN automatically benchmarks several modes of operation for primitive functions in run-time, which might change between runs. 
    
    %%% Processing unit
    \textbf{Processing unit} Changing the processor can affect results \cite{hong_2013,gundersen_2022}. 
    Nagarajan et al. \shortcite{nagarajan_2019} found that a deterministic GPU implementation repeatedly generated the same result when executed on the same GPU, but changed to a different, but deterministic result, when executed on another GPU.

    %%% Rounding errors
    \textbf{Rounding errors} Different hardware architectures and software implement the rounding of floating-point numbers in different ways, the rounding errors accumulate during long running calculations, particularly when using GPUs \cite{taufer_2010}.

\textit{Implementation factors can cause outcomes to differ if software, hardware or initialization seeds are changed between experiment runs or parallel processing is utilized.}
Deterministic implementations of primitive operations, the use of single thread processes and forcing execution in serial manner will guarantee deterministic results \cite{pham_2020} - for a given setup. 
%Analysis can also be affected if any performance metrics rely on stochasticity such as sampling.

%%%%%%%%%%%%% SECTION - OBSERVATION FACTORS
\section{Observation Factors} 
Observational factors are related to how data is generated, processed and augmented, but also to the properties of environments used for benchmarking, such as agent simulation environments.

    %%% Dataset bias
    \textbf{Dataset bias} The methods used to gather data (manual or automated) and the way data is captured (objects are often centered when photographed) introduce biases in datasets \cite{torralba_2011}.
    Recth et al. \shortcite{recht_2019} show that algorithms generalize poorly even on datasets that are replicated using the same source populations, so the lack of access to data used in the original study could lead to differences in data distribution for reproducibility experiments \cite{pineau_2021}. %Documentation instead => Dataset not available?
    In the social and environmental sciences, models might not generalize from one geographic area to another because of spatial dependence and heterogeneity \cite{goodchild_2021}
    Dataset shifts is also an issue \cite{finlayson_2021}.
    \textit{Different datasets lead to different outcomes.}
    
    %%% Pre-processiong
    \textbf{Pre-processing} Differences in data pre-processing will change data samples, so the applied pre-processing techniques must be well documented to facilitate reproducibility \cite{dacrema_2021}.%,gundersen_2021a}. 
    \textit{ Differences in data pre-processing changes outcomes.}
    
    %%% Data augmentation
    \textbf{Data augmentation} Stochastic data augmentation procedures are influenced by both algorithmic and implementation factors, which leads to differences in training data and thus different outcomes \cite{pham_2020,zhuang_2021,bouthillier_2021}. % they do not discuss which methods they use, conclusion is that you should not seed the data augmentation (!).
    
    %%% Data splits
    \textbf{Data splits} Difference in data splits cause a difference in outcomes \cite{makridakis_2018}, which includes stochastic sampling from the training set instead of training on a static validation set \cite{bouthillier_2019,bouthillier_2021}.
    Also, random selection of samples during training is typically used in gradient boosted trees \cite{pouchard_2020}. 
    According to Gundersen and Kjensmo \shortcite{gundersen_2018}, only 16\% specify the validation set and 30\% specify the test set, which means that for all practical purposes outcome reproducibility is impossible to achieve. 
    A 2\% to 12\% variation has been shown in labeled attachment scores of Natural Language Processing experiments when comparing models trained using standard test splits versus random test splits, which suggests that the results reported by experiments that only use the standard test splits can be influenced by a bias in the standard test splits that favor certain types of parsers \cite{ccoltekin_2020}.

    %%% Environment properties
    \textbf{Environment properties} Stochasticity and different dynamic properties of the testing environment could affect the outcome, especially in continuous control simulators such as those used in deep reinforcement learning \cite{henderson_2018}.
    
    %%% Annotation quality
    \textbf{Annotation quality} Differences in annotations made by humans will affect the target value and thus the outcome a model produces \cite{belz_2021b}.
    
    %%% Test data issues
    \textbf{Test data issues} Data leakage results in models trained on data that should only be available at test time, leading to overestimating model performance \cite{dacrema_2021}. %Check reference 
    Götz-Han et al. \shortcite{gotz_2020} demonstrate five cases of reported performance gains well above the state-of-the-art that were the results of data leakage. 
    The performance gains are far below the claims of the original researchers when the data leakage errors were corrected.
    Cases where metrics have been reported on training data instead of test data has also been found \cite{kurach_2018}.
    Neither outcome nor analysis are changed, but the interpretation could lead to false conclusions . 
    
\textit{Observation factors might affect the outcome and interpretation of an experiment. 
The effect of these factors can be reduced by setting the random seed, sharing details about pre-processing and data provenance \cite{gebru_2021}.
} What is done with duplicate data, outliers and missing values can introduce biases \cite{stodden_2015}.
As long as datasets are finite samples of an infinite population \cite{melis_2017}, they might not reflect the actual distribution at a given point in time and can also shift over time.

%%%%%%%%%%%%% SECTION - EVALUATION FACTORS
\section{Evaluation factors}
Evaluation factors relate to how the investigators reach the conclusions from doing an experiment.
    
    %%% Selective reporting of results
    \textbf{Selective reporting results} through careful selection of datasets, and ignoring the danger of adaptive over-fitting could lead to the wrong conclusions being made \cite{pineau_2021}.
    %Motivation why one does stuff => Jannach?
    \textit{Selective reporting will affect the interpretation.}
    
    % Over-claiming of results
    \textbf{Over-claiming of results} By drawing conclusions that go beyond the evidence presented (e.g. insufficient number of experiments, mismatch between hypothesis and claim) results are over-estimated \cite{pineau_2021}.
    \textit{Over-claiming of results are errors in the interpretation.}

    %%% Lack of naive baselines
    \textbf{Lack of naïve baselines} Lack of comparison with simple statistical methods or naïve benchmarks such as linear regression and persistence in time-series forecasting could obscure results \cite{makridakis_2018}.
    \textit{Naïve baselines help interpret the performance, but will not affect outcome nor analysis.}

    %%% Sampled metrics
    \textbf{Sampled metrics} Sampling from the test set is used sometimes when evaluations are computationally demanding. This could lead to sampled metrics being inconsistent with the exact versions and thus the wrong conclusions can be inferred \cite{cremonesi_2021}.
    \textit{Sampled metrics will not change the outcome but can change the interpretation of the analysis.}
    
    % Error estimation
    \textbf{Error estimation} Machine learning methods must be able to specify certainty and confidence intervals around them \cite{makridakis_2018} and preferably testing statistical significance taking the confidence intervals into account \cite{henderson_2018}.
    Reporting single scores without any estimate of error or variation in performance is insufficient to compare non-deterministic approaches \cite{reimers_2017}.
    Error estimates are more prevalent in machine learning experiments reported in healthcare than for other domains %such as image recognition, natural language and general artificial intelligence 
    \cite{mcdermott_2021}.
    \textit{Error analyses are part of the analysis and could change the interpretation if done incorrectly or are lacking.}
    
    %%% Statistical analysis
    \textbf{Statistical analysis} Improper use of statistics to analyze results, such as claiming significance without proper statistical testing or using the wrong statistics test lead to false conclusions \cite{card_2020,pineau_2021}. 
    Power analyses should be done prior to evaluation when comparing against baselines; the number of instances in the test will determine the effect size and should be chosen accordingly \cite{card_2020}.
    \textit{The decision of which statistical analysis to do affects the analysis.}

\emph{Evaluation factors affect the analysis and the interpretation of the analysis and thus the conclusion.} 
Evaluation factors can only be controlled through validation and ensuring that one is doing the right experiment and evaluating it correctly.
Sculley et al. \shortcite{sculley_2018} list the following practices that should be included in empirical studies: 1) tuning methodology, 2) sliced analysis, 3) ablation studies, 4) sanity checks and counterfactuals and 5) at least one negative result. 
However, they note that a material increase in standards for empirical analysis or rigor has not been observed across the field.

%Historic data not the same as real-time production data (from laboratory to production paper referred to in the AI Mag Jannach paper.

%%%%%%%%%%%%% SECTION - DOCUMENTATION FACTORS
\section{Documentation Factors}
Documentation factors are related to how well an experiment is documented, which means \textit{ideally} documenting all the choices mentioned above, which can be impractical.
    
    %%% Readibility
    \textbf{Readability} 
    The readability of papers influences whether it is possible to reproduce results. 
    Mathiness could lead to reduced readability \cite{lipton_2018}.
    Gundersen and Kjensmo \shortcite{gundersen_2018} found that a large degree of papers only implicitly state what research questions they answer (94\%), which problems that they seek to solve (53\%), and what the objective (goal) of conducting the research is (78\%). 
    Only 5\% of the papers explicitly states the hypothesis and 54\% contain pseudo-code. 
    Raff \shortcite{raff_2019} found that number of tables, readability, specification of hyperparameters, pseudo-code, number of equations and compute needed to run the experiment correlated strongly with reproducibility.
    \textit{Readability could affect the outcome, analysis and interpretation.}
    
    %%% Experiment design details
    \textbf{Experiment design details}
    %Lack of availability of the code necessary to run the experiments \cite{pineau_2021,mcdermott_2021}.
    Under-specification of the metrics used to report results and misspecification or under-specification of the model or training procedure might lead to the wrong conclusions \cite{pineau_2021}.
    \textit{Documentation that lacks experiment details could affect outcomes, analyses and interpretations.}
    
    %%% Workflow
    \textbf{Workflow}
    The exact steps taken and their order when conducting the empirical machine learning studies will affect the outcome \cite{gundersen_2021a}, especially when they become more complex \cite{rupprecht_2020}. 
    Missing documentation of steps could also affect the reproducibility. 
    \citet{kurach_2018} found that data augmentation was not specified, even when data had been augmented.
    \textit{Not specifying the workflow properly could lead to different outcomes.}
        
    %%% Implementation details
    \textbf{Implementation details} 
    Details on how novel algorithms and baselines are implemented, especially details that can affect reproducibility, are important \cite{henderson_2018}. 
    Inconsistencies in the documentation and implementation of software can cause reproducibility experiments to fail when using different software. 
    \citet{alahmari_2020} showed how a Karas' documentation error stating that convolutional layers weights initialization was based on the Glorot uniform, however the actual implementation was with modified version of Glorot uniform, known as the Xavier uniform, causing the results to not be reproducible when implemented on PyTorch.    
    \textit{Lack of implementation details of the machine learning algorithms could lead to differing outcomes, while for performance metrics it could lead to different analysis.}
    
    %%% Access to data
    \textbf{Access to data}
    The availability of data is required for outcome reproducibility for non-trivial data, such as synthetic data that can be generated by rules.
    However, data cannot always be made publicly available \cite{pineau_2021} and might not be easily available to collect, i.e. medical data is sensitive \cite{mcdermott_2021}.
    \textit{Availability of data may affect the outcome.}
    % CK: changed conclusion, someone should validate
    
    %%% Access to code
    \textbf{Access to code}
    Code describes implementation details perfectly and is required for outcome reproducibility for experiments of some complexity.
    Reproducing the experiments will require more effort if the code that is necessary to run the experiments is not available \cite{pineau_2021,mcdermott_2021,gundersen_2019}. The version number or commit ID should be noted when referencing code in a git repository.
    \textit{Lack of code could both affect the outcome, analysis, and interpretation.}

    %%% Stale URLs
    \textbf{Stale URLs} URLs in papers that link to software and data often stop working. 
    \citet{hennessey_2013} analyzed 14,489 unique web pages found in the abstracts of papers published between 1996 and 2010 and found that the median lifespan of these web pages was 9.3 years with 62\% of them being archived.
    \emph{Can affect both outcome and analysis.}
    
Documentation factors affect all reproducibility degrees, but can be alleviated by sharing code and data since the code and data themselves document so many aspects of the experiment. 
This is why code and data sharing is so effective for increasing reproducibility. The location and stability of the resource should be considered when posting data and code to enable reproducible results.

\section{Discussion}

%%%
%%% RELATED WORK
%%%
%\section{Related Work}
%\label{sec:related_work}
In recent years, several studies have sought to investigate potential sources of irreproducibility in machine learning. % without relating these sources to a definition of reproducibility. 
Most studies have investigated what are called implementation factors and algorithmic factors by Pham et al. \shortcite{pham_2020} and Zhuang et al. \shortcite{zhuang_2021}, which both will lead to differing output between runs. 
Algorithmic factors are design choices related to randomness being introduced in different steps of  machine learning algorithms, such as initialization and feature selection. 
Implementation factors are design choices related to, for example, which seeds are used when initializing the pseudo random number generators, which software and software versions the experiments require or the hardware the experiments are executed on. 
%All of these can affect the exact output produced by the machine learning algorithms.
Other studies have investigated how the documentation of the experiments can be a source of irreproducible results 
\cite{gundersen_2018,raff_2019,raff_2021}.
Improper design of a study can also make it irreproducible \cite{simonsohn_2014,cremonesi_2021,dacrema_2021,cockburn_2020}.
Finally, studies have pointed out that the design and execution of the evaluation can affect results \cite{pineau_2021,makridakis_2018,card_2020}.

Despite reproducibility being a cornerstone of science, Plesser \shortcite{plesser_2018} argues that reproducibility is a confused term.
Many definitions exist in the literature, see both \cite{plesser_2018} and \cite{gundersen_2021a} for reviews, but none of these definitions are easy to operationalize; it is not straight forward for researchers or practitioners to let the definitions guide them when trying to reproduce research and analyze failures to reproduce. 
The definitions do not help in understanding what exactly is required of a reproducibility experiment nor how the results should be evaluated to be considered a success or a failure.
%They do not even characterize what entails a reproducibility experiment. 
We will exemplify this by reviewing some of the most relevant definitions of reproducibility. 

The Association for Computing Machinery (ACM) \shortcite{ACM_2020}, who based their definition on the Joint Committee for Guides in Metrology \shortcite{JCGM_2012}, defines reproducibility as "the main results of the paper have been obtained in a subsequent study by a person or team other than the authors, using, in part, artifacts provided by the author" while replication is defined as "the main results of the paper have been independently obtained in a subsequent study by a person or team other than the authors, without the use of author-supplied artifacts."
These definitions are open for interpretation.
It is not clear what exactly is meant by "main results have been obtained". 
Because of this, concluding whether a reproducibility experiment is a success or not is subjective.
Furthermore, except for in a replication where no artifacts should be  used, it is not clear which artifacts should or can be used for a reproducibility study.
However, the definition by ACM is not alone in this ambiguity.

The definitions proposed by The U.S. National Academy of Science \shortcite{NAS_2019} are also ambiguous.
They define reproducibility as \emph{"obtaining consistent computational results using the same input data, computational steps, methods, and code, and conditions of analysis"} and replicability to mean \emph{"obtaining consistent results across studies aimed at answering the same scientific question, each of which has obtained its own data."}
In a similar fashion as ACM, it is not clear how to interpret "obtaining consistent results" in a practical setting.
However, this definition clearly states what is the input to reproducibility studies and replications.
Replication differs in stating that only data needs to be different in a replication study, which means that computational steps, methods, code and conditions of analysis can be the same.
Still, it is not clear exactly how computational steps differs from code and methods, so even when being more precise the definitions are still ambiguous. 

Peng \shortcite{peng_2011} also distinguishes between replication and reproducibility. 
While replication requires new evidence (in the form of data) for scientific claims to be independently evaluated, reproducibility requires code and data to be published.
According to \citet{peng_2011}, the least reproducible research  requires code to be published so that independent researchers can review it.
Next level requires also data to be published while the gold standard is to share linked and executable code and data. 
Peng \shortcite{peng_2011} mentions that all the papers that were reviewed for reproducibility published in the journal \emph{Biostatistics} at the time were reproducible. 
However, it is not clear what was required of a paper to be evaluated as reproducible. 
The same is an issue with the definitions. 

Goodman et al. \shortcite{goodman_2016} do not distinguish between reproduciblility and replication. 
They view this to be the same concept, and the define three different reproducibility levels, results reproducibility, method reproducibility and inference reproducibility. 
These definitions have similar issues with ambiguity with regards to input of a reproducibility experiment and how to interpret the results. 
Gundersen \shortcite{gundersen_2021a} tries to solve this amiguity by proposing \emph{reproducibility types} specifying which documentation (paper, code and data) provided by the authors of the study that are being utilized by independent researchers in the reproducibility experiment as well as introducing \emph{reproducibility degrees} that specify how results can be interpreted to be considered a success.
Further details are given in section \ref{sec:model_comparison_and_reprodicibilty}.

%None of the definitions presented above, nor any other we are aware of, clarify which design choices exactly can lead to studies being irreproducible.
Reproducibility definitions are not easy to operationalize because of their inherent ambiguity.
No articles in literature, that we are aware of, provide an overview of experiment design choices that can lead to irreproducible results, nor do any articles try to relate the design choices to the interpretation of results. 
%This means that the guidelines and reproducibility checklists are not properly rooted in theory and machine learning methodology, but are made based on experience.
This article seeks to remedy these issues by providing a framework for reproducibility in machine learning that can easily be operationalized by researchers and practitioners to reduce failures of impossible tasks, engineering, post-deployment and communication \cite{raji_2022}. 
%First, machine learning methodology is introduced. 

According to \citet{ioannidis_2005}, the greater the flexibility in designs, definitions, outcomes, and analytical modes in a scientific field, the less likely the research findings are to be true.
This holds true for machine learning, as "the rate of empirical advancement may not have been matched by consistent increase in the level of empirical rigor across the field as a whole" as \citet{sculley_2018} phrase it.
Machine learning methodology is concerned with ensuring that models generalize well to unseen data. 
Information from the test set should not be used to optimize the performance of a model.
This is why machine learning competitions, such as the ImageNet Large Scale Visual Recognition Challenge (ILSVRC), restrict the number of submissions per week. 
Breaking this rule resulted in a ban at the 2015 challenge \cite{markoff_2015}.
Less attention has been given in machine learning methodology to how information can flow in a similar way from the  experiment conclusion to the experiment design. 
The conclusion should not change the protocol on how it is inferred. 
Information should not flow from the conclusion to the analysis and further change how the experiment is re-executed or analyzed.
This is especially important in the sciences, including computer science \cite{denning_1988} and general and machine learning more specifically \cite{russel_norvig_2020}, where the design of an artifact is as important as its experimental evaluation. %Reference?
Typically, algorithm design and experimental evaluation are done iteratively until progress has been made, which is a process that can be prone to methodological mishaps unless proper care is taken. 

The framework would be highly relevant for reproducibility studies. 
The detailed description of the reproducibility study conducted by \citet{nunzio_2023} did not rely on a reproducibility framework such as the one proposed here.
While thorough, some relevant details are lacking. 
The authors base the reproducibility study on the article, code and data provided by the original authors, so it is clearly of type R4 Experiment.
A difference in performance of almost 2 percentage points from the original study is reported.
However, the study lacks a systematic and thorough discussion on the potential sources to this difference. 
Also, a good discussion on the improper evaluation of results in the reproduced study does not contain potential consequences of the evaluation.
As performance differs, the reproducibility degree is not outcome reproducible (O) and the same methodology to evaluate the results is used so it is not interpretation reproducible (I). 
However, it is not clear whether the reproducibility study supports the conclusion of the original study and hence confirms the original study. 
If this was the case, the study would have been classified as \textit{R4A}.

\section{Limitations}
While we believe that the grouping of design decisions into six categories might be at the right level and be exhaustive, the overview and taxonomy do probably not contain all design decisions that can lead to irreproducible results. 
Our goal is to capture as many as possible, but some might have eluded us.
One way of being more certain about capturing as many design decisions as possible would be to perform a structured literature review.
We did not do this, but not from a lack of trying.
However, a search for "machine learning" AND "reproducibility" and similar terms returns too many irrelevant articles to be practically doable. 
Instead, we have relied on following the literature for several years.
Also, this article does not contain any experiments; it is an overview with references. 
We do not quantify the importance of the different factors nor the extent to which they can affect the outcome either. 
The extent to how much performance might vary have been evaluated by other studies cited here. 
Also, according to these studies the difference in performance differs between machine learning methods and even deep learning architectures, so there is no clear rule of thumb.
Hence, what is important is to test for variation and characterize it using statistical methods and through this increase rigor. 
Increasing rigor is exactly what we seek to support with this framework by showing how factors affect conclusions.
%If this is not done, even seemingly clear results might not be statistically significant. 

%Another important distinction that could be made would be between inter- and intra-laboratory reproducibility, as is done for example in analytical chemistry \cite{miller_2018}. 
%Inter-laboratory reproducibility is a measure of variance in results between different laboratories, while intra-laboratory reproducibility measures variance in the results within a single laboratory. A similar distinction can be made in machine learning if one considers differing hardware and software as inter-laboratory conditions. 

%A possible approach to change the current cost-benefit analysis framing is introducing reproducibility contributions from researchers other than the authors and the readers.  Many students start their research careers by reproducing published papers.  These efforts could be incorporated into the cost-benefit equation as those researchers could add documentation about experiments that augments what the authors provided. Their incentive could be getting a separate publication. 
% that documents thoroughly their own successful reproduction of the original work.  The overall documentation needed remain the same, but the effort would be required of the original authors would be greatly reduced as well as the effort to others to reproduce the work.  

%%%%%%%%%%%%%%%%%%%%%%%%%%%%%%%%%%%%%%%%%%%%%%%%%%%%%%%%%%%%%%%%%%
%%% CONCLUSION
%%%%%%%%%%%%%%%%%%%%%%%%%%%%%%%%%%%%%%%%%%%%%%%%%%%%%%%%%%%%%%%%%%
\section{Conclusion}
The main contribution of this paper is the identification and categorization of 41 design choices documented in the literature that can lead to false conclusions.
Another major contribution is a novel framework that enables applied data science researchers and practitioners to understand which experiment design choices can lead to false findings, understand how these design choices can affect the conclusion of experiments and conduct and analyze reproducibility experiments.
It is the first comprehensive framework for machine learning reproducibility that provides an overview and characterization of factors that affect reproducibility in machine learning experiments, and it extends existing reproducibility checklists for authors by identifying additional factors that span all levels of the technical stack including hardware.  
This is also the first work to describe how the reproducibility factors affect the conclusions that are drawn from experiments by relating those factors to the scientific method and different definitions and scope of reproducibility studies, 
In an era where reproducibility is a priority in all areas of science, our goal is to shed light on the reproducibility challenges and needs in machine learning so that forward-looking solutions and methodologies can stem from this discipline and lead the way for other communities.  

\bibliographystyle{ACM-Reference-Format}
\bibliography{kdd2023}

\end{document}